# Learning Regular Expressions for Interpretable Medical Text Classification

Chaofan Tu
*School of Computer Science*
*University of Nottingham Ningbo China*
Ningbo, China
chaofan.tu@nottingham.edu.cn

Ruibin Bai
*School of Computer Science*
*University of Nottingham Ningbo China*
Ningbo, China
ruibin.bai@nottingham.edu.cn

Zheng Lu
*School of Computer Science*
*University of Nottingham Ningbo China*
Ningbo, China
zheng.lu@nottingham.edu.cn

Uwe Aickelin
*School of Computer and Information Systems*
*University of Melbourne*
Melbourne, Australia
uwe.aickelin@unimelb.edu.au

Peiming Ge
*Technology Department*
*Ping An Health Cloud Company Limited*
China
Shanghai, China
gepeiming649@jk.cn

Jianshuang Zhao
*Technology Department*
*Ping An Health Cloud Company Limited*
China
Shanghai, China
zhaojianshuang059@jk.cn

*Abstract*—we considered a text classification system that uses auto-generated regular expressions for high quality and fully-interpretable medical use. Text classification is an area that has been extensively studied for a long time, and many different approaches have been studied. In this paper an algorithm called PSAW is proposed for learning regular expressions, which combines pool-based simulated annealing with word vector model to satisfy the requirement of interpretability and readability in medical filed. Tested on 30 medical text classification tasks, each containing half a million real-life data from one of China's largest online medical platform, PSAW has shown much potential compared to domain experts and most of those classifiers by PSAW are human readable for further revisions and validation.

## I. INTRODUCTION

With the prevalence of modern computerised technologies, the chances to boost the accuracy of the auxiliary diagnosis models are manyfold which in turn enhances the doctor/hospital operational efficiency. By using latest technologies in speech recognition, machine vision, Natural Language Processing (NLP), machine learning, and others, data mining is becoming a necessity in the field of intelligent healthcare sector where huge amount of digital data is available. For instance, the IBM Watson for ontology has demonstrated concordance rates of 96% for lung, 81% for colon, and 93% for rectal cancer cases with a multi-disciplinary tumour board in India [1]. The Watson collects the data from 8500 hospitals, insurers, and government agencies [2]. Another popular application of intelligent healthcare is the DeepMind's Streams medical diagnosis mobile application. It sends nurses and doctors alerts when a patient's readings look abnormal through analysing the medical records of 1.6 million patients [3]. The availability of vast quantity of digitised healthcare and patient data plays an important role in the auxiliary diagnosis models.

This present work is based on the cooperation with a very large online medical platform,. The average number of daily consultation requests submitted to this platform in 2017 exceeded 370,000 [4]. In order to assign appropriate doctors across the different disciplines such as gynaecology, paediatrics, dermatology and so on, the system has to first deal with the classification of medical inquiries, which are sentence-level patients' complaints.

Any mistakes in the classification process will lead to doctor miss-assignment and therefore reduce the system overall efficiency, especially in the real-time online scenarios. For the sake of not only reducing the number of employees who should be available 24/7 handling the medical reception but also enhancing the platform's operational efficiency, it is very important to implement an automated classification system. Unlike traditional classifiers, classifiers intended for medical text need to be with good interpretability and readability, due to the rigorous validation requirements of the medical filed.

Regular expressions are widely used in text matching techniques which are fully interpretable compared to deep learning models. This paper proposes a fully-automated system for regular expressions to solve medical text classification problems. The contributions of this paper are as follows:

- A specially designed structure of solutions is proposed to reduce the complexity whilst maintaining flexibility;
- An algorithm called PSAW is proposed, combining a pool-based simulated annealing and the word-vector model to enhance the readability of auto-generated regular expressions.
- Impacts of parallel and iterative strategies for tasks of learning regular expressions have been intensively explored by comparing two extended versions of PSAW.

## II. RELATED WORKS

Text Classification involves assigning a text document to a set of pre-defined classes automatically. Classification is usually done on the basis of significant words or features extracted from the raw textual document. Since the classes are pre-defined, it is a typical supervised machine learning task [5]. Automated text classification usually includes steps such as pre-processing (eliminating stop-words, etc.), feature selection using various statistical or semantic approaches, and text modelling [5]. Until late 80's, text classification task was based on Knowledge Engineering (KE), where a set of rules were defined manually to encode the expert knowledge on how to classify the documents given the categories [6]. Since there is a requirement of human intervention in knowledge engineering, researchers in 90's have proposed many machine learning techniques to automatically manage and organise the textual documents [6]. The advantages of a machine learning based approach are that the accuracy is comparable to that of human experts and no artificial labour work from either knowledge engineers or domain experts needed for the construction of a document management tool [7].

Text classification involves challenges and difficulties. First, it is difficult to capture high-level semantics and abstract concepts of natural languages just from a few key words. Second, semantic analysis, a major step in designing an information retrieval system, is not well understood. Third, high dimensionality (thousands of feature vectors) of data poses negative influences for classification tasks [8].

Before text classification, text representation is the first problem. Bag of Words (BoWs) is one of the basic methods of representing a document. The BoWs is used to form a vector representing a document using the frequency count of each term in the document based on a fixed global vocabulary. This method of text representation is called as a Vector Space Model (VSM) [9]. Unfortunately, BoWs/VSM representation scheme has its own limitations. For example, high dimensionality of the representation, loss of correlation with adjacent words, and absence of semantic relationship [10]. Another VSM-based method is a neural network based model called Word2vec which is used in this paper for distributed word embeddings, which was proposed by Mikolov et al. in 2013 [11, 12]. The fixed length vector representation trained by word2vec deep learning model has been shown to carry semantic meanings and are useful in various NLP tasks such as text classification, speech recognition, and image caption generation [13].

After text presentation, word embeddings or numerical representations for feature extraction of texts can be fed into plain classifiers like the Naïve Bayes, decision tree, neural network, support vector machine, hybrid approaches etc. [8]. The Naïve Bayes classifier is the simplest probabilistic classifier used to classify the text documents into predefined labels [8]. The Nearest Neighbour classification is a non-parametric method and it can be shown that for large datasets the error rate of the 1-Nearest Neighbour classifier is not likely to be larger than twice the optimal error rate [8]. Centroid based classifier is a popular supervised approach used to classify texts into a set of predefined classes with relatively low computation [8]. Decision trees are the most widely used inductive learning methods [8]. Decision trees' robustness to noisy data and their capability to learn disjunctive expressions seem suitable for document classification[8]. A Support Vector Machine (SVM) is a supervised classification algorithm that has been extensively and successfully used for text classification tasks [8]. Neural Network based text classifier are also prevalent in the literature, where the input units are denoted as feature terms, the output unit(s) are the category or categories of interest, and the weights on the edges connecting units form dependence relations [8].

A series of experiments of sentence-level classification tasks with a simple convolutional neural network (CNN) built on top of word vector model suggest that unsupervised pre-training of word vectors is an important ingredient in deep learning for NLP [14]. Neural network based approaches are strong alternatives but usually less interpretable because those black box models cannot be logically explained [15]. In addition, those black box models cannot be quickly modified except retraining models [16]. To address those difficult issues discussed above, some related work has been done by using regular expressions for classification tasks, and some auto-generated regular expressions can be effectively used to solve the classification problems as an interpretable way.

A novel regular expression discovery (RED) algorithm and two text classifiers based on RED were designed to automate both the creation and utilisation of regular expressions in text classification [17]. The proposed RED+ALIGN method correctly classifies many instances that were misclassified by the SVM classifier. A novel transformation-based algorithm, called ReLIE, was developed for learning such complex character-level regular expressions for entity extraction tasks. The experiments demonstrate that it is effective for certain classes of entity extraction [18].

Automated regular expressions learning can also be viewed as a data-driven optimisation problem. In this paper, a well known simulated annealing hyper-heuristic [19] has been adapted for learning regular expressions for text classification. The choice of this approach is based on the fact that there are naturally multiple neighborhood operators available for generating regular expression variants and hyper-heuristics can learn to orchestrate the selections of different operators to achieve high performance across different problems. It has been shown that specially designed neighborhood operators of SA will lead to better performance [15].

## III. PROBLEM DESCRIPTION

Formally the problem can be defined as follows: given a set of predefined classes $C$ (or medical templates in the context of our application) and a set of text inquires $Q$, the problem is to classify each inquiry $q \in Q$ to one of classes $c \in C$ automatically based on a set of previously labelled examples by medical experts. Table I shows examples of the classification, text inquiry is usually a piece of text information given by the user, describing the medical conditions or problems; the classification task is to select the most appropriate medical template for this inquiry.

TABLE I. EXAMPLES OF MEDICAL TEXT CLASSIFICATION

| Text inquiry | Medical template |
|---|---|
| "My girl is three years old and always coughs without fever, what can I do for her?" | Cough: 1-3 years old child |
| "I have been suffered from pain in my lower abdomen for 3 weeks." | Adult bellyache |
| "The acne grows on the back, and recently it is a little itchy near it." | Folliculitis |
| "I have a serious hair loss, how to make hair?" | Hair loss |

Let $R$ be a regular expression designed for classification of class (or medical template in our application) $C$ (denote $|C|$ be the number of inquiries in class $C$), and let $M(R, Q) \in Q$ be the set of all medical texts matched by $R$ in $Q$; Denote $M_p(R, Q) = \{q \in M(R, Q): q$ is an instance of $C\}$ to the set of all correctly matched entries (medical text inquires) and denote $M_n(R, Q) = \{q \in M(R, Q): q$ is not an instance of $C\}$ is the set of all mismatched entries (medical text inquires). Like many classification problems, this problem also has two performance indicators, which are precision and recall as bellows:

$$precsion(R,Q) = \frac{|M_p(R,Q)|}{|M_p(R,Q)|+|M_n(R,Q)|} \quad (1)$$

$$recall(R,Q) = \frac{|M_p(R,Q)|}{|C|} \quad (2)$$

The well-known F-measure (also called F-score) can be a better single metric when compared to precision and recall. Given a non-negative real $\beta$ for users' preference, it can be expressed as:

$$F_\beta(R,Q) = \frac{(1+\beta^2) \times precsion(R,Q) \times recall(R,Q)}{\beta^2 \times precison(R,Q) \times recall(R,Q)} \quad (3)$$

The problem of automated learning of classifiers for medical text in this paper can be formally expressed as an optimization problem for regular expression $R$. Let $S$ be the solution space of $R$, for a given class of $C$ and labelled dataset $W$ which can be divided into a positive part and a negative part, the problem is to find a solution with the optimal objective function F-measure from the solution space $S$. So this problem can be defined as:

$$R_{target} = argmax_{R \in S} F_\beta(R,Q) \quad (4)$$

## IV. METHODOLOGY

### A. Proposed Structure

In this problem, each solution is encoded as a vector of $m$ regular expressions $< R_1, R_2, ..., R_m >$. To check whether a particular inquiry belongs to a class (or template), the regular expressions in the vector is executed one by one sequentially in the same order of the vector for the inquiry under consideration. If the inquiry is matched by any of regular expressions, the inquiry is said to be in the class, otherwise it is not in the given class. Each regular expression $R_i$ is derived via a combination of functions and terminals defined in Table I and follows a global structure of two parts $P_i$ and $N_i$ concatenated by the *NOT* function #_#, where $P_i$ tries to match all positive inquiries and $N_i$ is then used to filter out the list of falsely matched inquiries by $P_i$. That is, each regular expression has the following format:

$$R_i = (P_i).(\#\_\#(N_i))$$

TABLE II. FUNCTIONS AND TERMINALS

| Name | Label | Description |
|---|---|---|
| NOT | #_# | Function to negate a given expression. |
| Expression | e | An expression or term obtained through a combination of words and functions listed below. |
| OR | / | Function to test logic or of two expressions. |
| Words | w | List of key words extracted from the target text set. |
| AND | . | Function to test logic AND of two expressions. |
| Adjacency | {a, b} | Function to test the whether the distance between two words $w_1$ and $w_2$ are in the range [a, b], which is an extended function of the AND function. |

For the purpose of better readability of regular expressions and reduced search space, the following constraints are also applied to each of the regular expressions $R_i$:

*1) Each regular expression $R_i$ has at most one NOT function.*

*2) The positive part $P_i$ and negative part $N_i$ are only composed of OR function which is defined as the outer OR structure as below:*

$$R_i = (e_{p1}/e_{p2}/...|e_{pm}).(\#\_\#(e_{n1}/e_{n2}/...|e_{nn}))$$

*3) Function OR in the sub-expressions which is defined as the inner OR structure should not contain any other nested functions except itself.* That is, expression $w_1/(w_2/w_3)$ is acceptable but expression $w_1|(w_2 \cdot w_3)$ is not permitted.

*4) Function AND in the sub-expressions can contain nested functions of both AND and OR.* For example, both expressions $w_1 \cdot (w_2 \cdot w_3)$ and $w_1 \cdot (w_2|w_3)$ are acceptable but $w_1|(w_2 \cdot w_3)$ is not permitted because it violates condition 3.

The outer OR structure is used to compose the positive part $P_i$ and negative part $N_i$ directly according to condition 2, while the inner OR structure is in the sub-expressions $e$ cannot only contain any nested function except itself due to condition 3, So the overall structure of regular expression $R_i$ has been limited to a maximum of two levels of nested OR structure through the above restrictions.

**Corollary 1** With the same terminals and functions listed in Table II, there always exists one or more regular expressions that satisfies all the above conditions and is equivalent to any expression without these conditions.

**Proof:** For condition 1, it's obviously because the *NOT* function is essentially one kind of set operation, multiple *NOT* functions can be reduced to one finally;

For condition 2, it's evident since the *OR* function has a lower priority than any other function except *NOT*, any other single or multiple functions can apply *OR* function to the outer layer of itself;

For condition 3 and 4, if an expression $e$ is directly composed of *OR* function and the *OR* function contains an *AND* function, $P_i$ can be clearly transform into a new expression which meets the condition 2 as below:

$$\text{Let } e_{p2} = w_1/(w_2 \cdot w_3),$$
$$P_i = (e_{p1}/(w_1/w_2 \cdot w_3)|...|e_{pm}) = (e_{p1}/w_1/w_2 \cdot w_3|...|e_{pm});$$

If an expression $e$ is composed of one *AND* function and the inner layer *OR* structure contains an *AND* function such as $(w_1/(w_2 \cdot w_3)) \cdot w_4$, $P_i$ can be also transform into a new expression which meets the condition 2 and 4 as below:

$$\text{Let } e_{p2} = (w_1/(w_2 \cdot w_3)) \cdot w_4,$$
$$P_i = (e_{p1}/((w_1/(w_2 \cdot w_3)) \cdot w_4)|...|e_{pm}) =$$
$$(e_{p1}/w_1 \cdot w_4|(w_2 \cdot w_3) \cdot w_4|...|e_{pm})$$

End of proof. That is, although we restrict the possible formats of our regular expressions to two-layer nested structures, their expressiveness are not reduced. These conditions not only simplify the structure of solutions, but also contribute to enhancing the readability and interpretability.

### B. Solution Pool Mechanism

According to problem description and the structure defined above, the medical text classification problem in this paper is transformed into a combinatorial optimisation problem. The simulated annealing algorithm is a large-scale combined problem global optimisation algorithm, which is widely used to solve the NP-hard combination optimisation problem.

In this paper, the simulated annealing algorithm is applied as the evolutionary computation algorithm, and a solution pool mechanism is designed and implemented to enhance the diversity of the solution, as shown in figure 1:

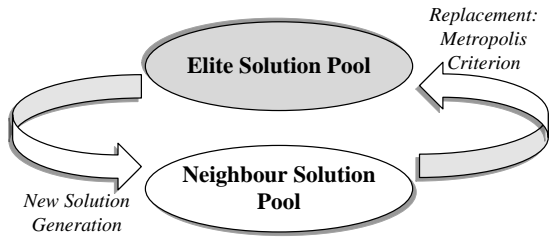

Fig. 1. Solution pool mechanism

The number of solutions in the elite solution pool is set to a fixed value, and the same amount of new solutions transformed from the initial solution are used as the initialization of the elite solution pool.

The number of solutions in the neighbour solution pool is the same as the elite solution pool. In each iteration of the entire period, each solution of elite solution pool produces a new solution, and all newly-generated solutions form a totally updated neighbour solution pool.

For each solution in neighbour solution pool, one in elite solution pool will be randomly selected for comparison and update. The acceptance criterion for solution replacement adopts metropolis criterion based on simulated annealing algorithm.

The best solution in the elite solution pool is always retained during whole period. The details of the proposed solution pool mechanism are shown as below in figure 2.

---
**Set** the capacity of solution pool to be $N_{pool}$;
**Define** a set of each elite solution $S_{e\_i}$ ($i = 1,..., N_{pool}$) as the elite solution pool $P_e$;
**Define** a set of each neighbour solution $S_{n\_j}$ ($j = 1,..., N_{pool}$) as the neighbour solution pool $P_n$;
**Set** the best solution in $P_e$ as $S_{e\_best}$;

**Solution Replacement:**
**begin**
    $j = 1$
    **while** ($j < N_{pool}$) **do**
        select $S_{n\_j}$;
        **select** $S_{e\_i}$ randomly from $P_e$;
        **let** $\delta$ be the difference in the evaluation function between $S_{e\_i}$ and $S_{n\_j}$;
        **if** the Metropolis criterion of simulated annealing is satisfied by $\delta$    $S_{e\_i} = S_{n\_j}$;   **endif**
        $j$ ++;

    **add** $S_{e\_best}$ into $P_e$ temporarily;
    **sort** $P_e$ by the evaluation function of each solution $S_{e\_i}$ ($i = 1,..., N_{pool}$ +1);
    $P_e = \{ S_{e\_i} \mid i = 1,..., N_{pool} \}$
    $S_{e\_best} = S_{e\_1}$
**end**

---
Fig. 2. Pseudo-code of the proposed solution pool mechanism

### C. Initialisation

The initial solution is a precondition for initialisation of the elite solution pool. In order to balance speed and readability, we carry out a method of word frequency and similarity comparison to generate a group of key words as the initial solution quickly. The specific steps are described as below:

*1) If the frequency of a word in the positive dataset exceeds a predetermined $TD_F$ times the frequency in the negative dataset, the word is added into the set of keywords;*

*2) Sorting the keyword set with the word frequency, and for the first predetermined $N_W$ keywords, calculating the cosine similarity between the two words' vectors, if it exceeds the predetermined $TD_S$, these two words are considered as two of a group of same subject words;*

*3) A group of subject words are randomly selected and connected as an inner OR structure as an initial regular expression without negative part.* Below is an exsample.

*((headache | dizzy | giddy | dizziness)).(#_#())*

*4) The single or multiple initial regular expressions generated by the step 3) form the initial classifier as the initial solution $S_{init}$.*

### D. Neighbourhood Operators

In this paper, it is decided by a random strategy whether to update the positive or negative part of regular expressions firstly, and then 7 specially designed neighbourhood operators are used for new solution generation.

O1: **Adding *OR* type 1** is an operator to add a word to the inner *OR* structure. First randomly select 10 words from the set of positive words or the set of negative words. Then randomly select an existed word from the inner nested *OR* structure, and calculate the similarity $Sim_i$ ($i = 1,..., 10$) between the existed word and the 10 words based on cosine similarity of pre-training word vectors. Finally, choose one word for adding from the 10 words based on probability:

$$Prob_i = \frac{Sim_i}{\sum_1^{10} Sim_i} \quad (5)$$

The extension of the inner *OR* structure combines the information of the pre-trained deep learning model of word2vec model, so that the readability of regular expressions has been considered during the evolutionary process.

O2: **Adding *OR* type 2** is an operator to add a sub-expression to the outer *OR structure*. First randomly select a word from the positive or negative word set. If the selected word does not exist in the outer *OR* structure, add the word; if the word is already in the outer *OR* structure, then randomly select another word to form a non-repeating *AND* (or *Adjacency*) sub-expression to add into the outer *OR* structure.

O3: **Removing *OR*** is an operator to randomly delete a sub-expression that makes up the outer or inner *OR* structure as an inverse operation of O1 and O2.

O4: **Adding *AND*** is an operator to extend the *AND* (or *Adjacency*) structure in the sub-expression in the outer *OR* structure. Randomly pick a word to insert into an existing *AND* (or *Adjacency*) structure or construct a new *AND* structure with a certain existing word.

O5: **Swap** is an operator to exchange the positions of any two sub-expressions in the *AND* (or *Adjacency*) structure.

O6: **Distance** is an operator to randomly change the maximally permitted distances between two expressions based on a given ***Distance Table***. Here the *AND* can be considered to be an *Adjacency* structure with unrestricted distance.

O7: **Removing *AND*** is an operator to randomly delete one sub-expression that makes up one *AND* (or *Adjacency*) structure, as an inverse operation of O4.

*E. Solution Decoding and Evaluation*

Each regular expression $R_i$ in solution should be decoded to a valid regular expression that can be passed through the general regular expression matching engine. There are two main points to note here. The logical symbols defined in this paper are not exactly the same as the symbolic system of regular expressions; The *NOT* function defined in this paper does not exist in regular expressions, so the positive and negative parts of $R_i$ need to be handled separately. A converted example is as follows:

Positive part of the converted $R_i$:
.*((($w_1$/$w_2$/$w_3$).*$w_4$/$w_5$.{0,10}($w_6$/$w_7$/$w_8$)/$w_9$/$w_{10}$|...|$w_p$)).* ;

Negative part of the converted $R_i$ :
.*((($w_1$/$w_2$/$w_3$/$w_4$/($w_5$/$w_6$).{0,10}$w_7$/$w_8$|...|$w_n$)).*

The performance of each solution will be evaluated based on the F-measure value according to the above description in section 2. The parameter $\beta$ for F-measure is set to 0.2 for the purpose of giving more attention to precision.

*F. Overall Algorithm*

This paper proposes a pool-based simulated annealing optimisation algorithm with Word2vec (PSAW), which is designed for automated learning of regular expressions to construct fully interpretable medical text classifiers. Below figure is the overall flow of PSAW.

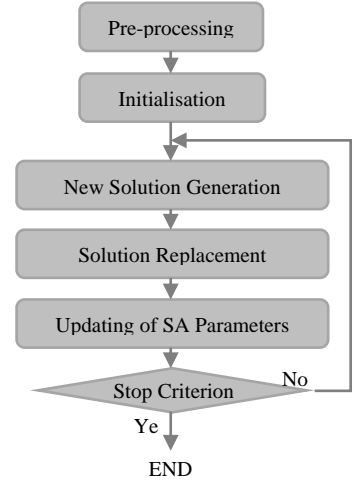

Fig. 3.    Overall process of PSAW

The pre-processing before initialisation includes dividing the training data set into positive and negative sets, performing Chinese word segmentation, removing stop words, and pre-training the Word2vec model.

Initial solution has been generated after the process of initialisation. At beginning the elite solution pool will be filled with new solutions from initial solution.

Solutions in elite solution pool may be replaced according to Metropolis criterion by solutions in neighbour solution pool and parameters such as the temperatures of the SA will be updated every iteration. The program terminates when the number of total iterations is over or the stop condition is met.

*G. Iterative and Parallel Strategies*

To further explore the impact of different operational strategies on time and performance, we designed and implemented an extended version of PSAW-I with iterative strategy and an extended version of PSAW-P with parallel strategy.

- PSAW-I: It is considered to improve the recall by iterative learning strategy, that is, before the learning of the next RE-based classifier, those entries matched by previous RE-classifiers in the training set should be filtered.

- PSAW-P: Consider the method of pre-dividing the positive training dataset for parallel acceleration. When the last parallel task is terminated, all the sub-solutions are merged as a solution for the whole. Pre-dividing is based on semantic clustering on our datasets and also set random division as comparison baseline.

## V. EXPERIMENT

### A. Data and Parameter Settings

Because of the collaboration with a very large online healthcare platform, the experiments in this paper are based on high quality training and test data from the real production environment. The numbers of real text inquiries in training set and test set are 1,800,000 and 500,000, respectively. The experiments in this paper used the following parameters, if not stated separately.

- For initialization: $TD_F = 5$; $N_W = 100$; $TD_S = 0.75$;
- For neighbourhood operators: *Distance Table* = [0, 2, 4, 6, 8, 10, 12, 14, 16, 18, 20, 100] ;
- For simulated annealing: starting temperature $T_S = 0.5$; stopping temperature $T_E = 0.05$; solution pool capacity $N_{pool} = 10$ ; Total iteration $K = 1000$ ;
- For F-measure: $\beta = 0.2$.

### B. Evaluation of Solution Pool Mechanism

In this experiment, we controlled the variables $N_{pool}$ and $K$ to learning one regular expression for a same template $C_1$ to evaluate the solution pool mechanism. The $N_{pool}$ of groups 1~3 were set to 1, 10, 50 and the $N_{pool}$ of group 4 was set to 1 to represent the traditional ways without this mechanism while the number of total new solution generated for group 4 was set as the same as group 2.

In TABLE III the results of groups 1~3 show that the higher the $N_{pool}$, the more time cost and the better the performance of F-measure ($F_m$).

The comparison of group 4 and group 2 shows that solution pool mechanism significantly not only enhances performance and but also reduces time cost. It is because without this mechanism the evolution will make the solution more and more complex to increase the evaluation time.

TABLE III. A COMPARISON OF POOL CAPACITIES

| Group | $N_{pool}$ | K | $F_m$ | Time (min) |
|---|---|---|---|---|
| 1 | 1 | 1000 | 0.60 | 8 |
| 2 | 10 | 1000 | 0.76 | 59 |
| 3 | 50 | 1000 | 0.77 | 364 |
| 4 | 1 | 10000 | 0.72 | 130 |

### C. Comparison of Iterative and Parallel Strategies

We have tested the PSAW algorithm and its two extended versions of PSAW-I, PSAW-P for six separate medical text classes $C_1 \sim C_6$. For further exploration, PSAW-P version adapted two division methods of clustering division and random division. All solutions here were set to contain 3 regular expressions, so one PSAW-P group was set to use the widely-used k-means clustering methods to divide the training dataset into 3 different parts while another PSAW-P group was set to use random trisection method for parallel processing.

The results in TABLE IV show that the PSAW-P version with clustering method shows the highest level of precision and the least time cost; the PSAW-I version with iterative strategy shows the highest level of recall, The PSAW original version itself shows the most time cost, while its average of F-measure value is the best.

TABLE IV. A COMPARISON OF PSAW WITH ITERATIVE AND PARALLEL STRATEGIES

| Classes | PSAW | | | | PSAW-I | | | | PSAW-P (Clustering) | | | | PSAW-P (Random) | | | |
|---|---|---|---|---|---|---|---|---|---|---|---|---|---|---|---|---|
| | Precision | Recall | $F_m$ | Time (min) | Precision | Recall | $F_m$ | Time (min) | Precision | Recall | $F_m$ | Time (min) | Precision | Recall | $F_m$ | Time (min) |
| $C_1$ | **0.89** | 0.63 | **0.87** | 266 | 0.76 | **0.76** | 0.76 | 220 | 0.87 | 0.42 | 0.83 | **86** | 0.86 | 0.52 | 0.84 | 110 |
| $C_2$ | 0.69 | 0.41 | **0.68** | 335 | 0.56 | **0.50** | 0.56 | 300 | **0.72** | 0.29 | 0.68 | **119** | 0.62 | 0.26 | 0.59 | 125 |
| $C_3$ | **0.71** | 0.11 | 0.58 | 417 | 0.52 | **0.25** | 0.50 | 372 | **0.71** | 0.11 | **0.59** | 138 | 0.66 | 0.09 | 0.54 | 144 |
| $C_4$ | **0.93** | 0.33 | **0.87** | 394 | 0.81 | **0.78** | 0.81 | 333 | 0.92 | 0.37 | **0.87** | 133 | 0.92 | 0.31 | 0.86 | 128 |
| $C_5$ | 0.84 | 0.69 | 0.83 | 405 | 0.87 | **0.83** | 0.87 | 345 | **0.89** | 0.54 | **0.87** | 136 | 0.86 | 0.51 | 0.84 | 134 |
| $C_6$ | **0.93** | 0.61 | **0.91** | 396 | 0.84 | **0.65** | 0.83 | 239 | **0.93** | 0.48 | 0.89 | 108 | 0.92 | 0.48 | 0.89 | 124 |
| **AVG** | 0.83 | 0.46 | **0.81** | 369 | 0.73 | **0.63** | 0.72 | 302 | **0.84** | 0.37 | 0.80 | **120** | 0.81 | 0.36 | 0.77 | 128 |

### D. Performance Distribution

The PSAW algorithm has been further applied to learning regular expression based classifiers for 30 independent disease templates to evaluate more of its performances compared to domain experts.

The followings figures 4 and 5 are precision and recall distribution of auto-generated classifiers by PSAW on the test dataset, compared to those manual classifiers written by domain experts.

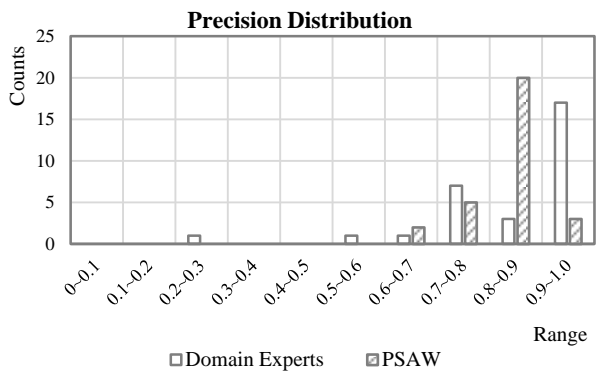

Fig. 4. Precision Distribution

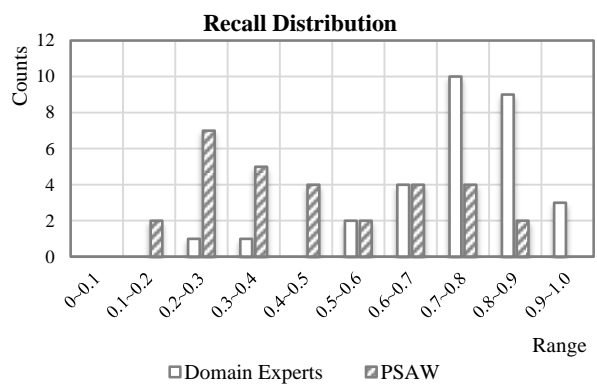

Fig. 5. Recall Distribution

Most precisions of classifiers by domain experts exceed 0.9 and auto-generated classifiers mostly exceed 0.8;

Most recalls of classifiers by domain experts exceed 0.6, while the distribution of auto-generated classifiers is more uniform. The reason may be the evaluation function we used was set to pay more attention to precisions rather than recalls ($\beta = 0.2$).

*E. Practicality Evaluation*

50 manual classifiers and 50 PSAW classifiers are randomly selected respectively for third-party practicality blind evaluation using a score table below. The distribution of results is shown in Figure 6.

TABLE V.  SCORE TABLE

| Score | Descriptions |
|---|---|
| 1 | Cannot be used |
| 2 | Can be used after a lot of revisions |
| 3 | Can be used after some revisions |
| 4 | Can be used after minor revisions |
| 5 | Can be used directly without revisions |

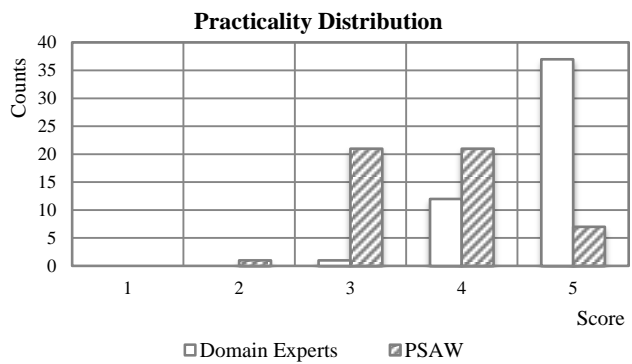

Fig. 6. Third-party blind evaluation

According to the third-party blind evaluation, most auto-generated classifiers by PSAW are well readable and can be applied to practical use after some or minor revisions, which benefit from the structure of solution and the use of word vector model.

## VI. CONCLUSION AND FUTURE RESEARCH

In this work, the medical text classification problem is transformed into a combinatorial optimisation problem. The proposed PSAW algorithm combines the classical simulated annealing with word vector model (pre-trained word2vec model) and has shown good potential compared to domain experts. Although those auto-generated classifiers by PSAW cannot outperformed experts' classifiers on each circumstance totally, most of them are fully interpretable and well readable for further revision to meet medial field's requirement. In addition, iterative and parallel strategies have been explored for further improvement on time cost and performance in this paper. Due to the good performance of PSAW, our partner has already applied this system for reducing labour work and accelerating the generation of regular expressions for practical use.

Future research includes using GPU to further speed up the algorithm, adoption of a multi-objective optimization model for higher level of recall, and more theoretical analysis research for more efficient regular expression encoding in the context of medical text classifications.

Acknowledgement: This work is partly supported by the National Natural Science Foundation of China (Grant No. 71471092), Natural Science Foundation of Zhejiang Province (Grant No. LR17G010001) and Ningbo Municipal Bureau of Science (Grant No. 2014A35006, 2017D10034). and Technology (Grant No. 2014A35006, 2017D10034).